\documentclass[11pt]{article}
\usepackage{scl}
\usepackage{times}
\usepackage{url}
\usepackage{latexsym}
\usepackage{lineno}
\usepackage{caption}
\usepackage{subcaption}
\usepackage{amsmath}
\usepackage[utf8]{inputenc}

\usepackage{graphicx}
\usepackage{array}

\title{Anveshana: A New Benchmark Dataset for Cross-Lingual Information Retrieval On English Queries and Sanskrit Documents}

\author{
    Manoj Balaji Jagadeeshan,
    Prince Raj,
    Pawan Goyal \\
    \texttt{Indian Institute of Technology, Kharagpur} \\
    \texttt{\{manojbalaji1, prfeynman0\}@gmail.com} \\
    \texttt{pawang@cse.iitkgp.ac.in}
}

\date{}

\begin{document}
\maketitle
%\linenumbers
\begin{abstract}
The study presents a comprehensive benchmark for retrieving Sanskrit documents using English queries, focusing on the chapters of the Srimadbhagavatam. It employs a tripartite approach: Direct Retrieval (DR), Translation-based Retrieval (DT), and Query Translation (QT), utilizing shared embedding spaces and advanced translation methods to enhance retrieval systems in a RAG framework. The study fine-tunes state-of-the-art models for Sanskrit's linguistic nuances, evaluating models such as BM25, REPLUG, mDPR, ColBERT, Contriever, and GPT-2. It adapts summarization techniques for Sanskrit documents to improve QA processing. Evaluation shows DT methods outperform DR and QT in handling the cross-lingual challenges of ancient texts, improving accessibility and understanding. A dataset of 3,400 English-Sanskrit query-document pairs underpins the study, aiming to preserve Sanskrit scriptures and share their philosophical importance widely. Our dataset is publicly available at https://huggingface.co/datasets/manojbalaji1/anveshana
\end{abstract}

\section{Introduction}
Sanskrit, a treasure trove of profound wisdom and philosophical insights, holds a significant place in India's cultural and spiritual heritage. This classical language includes a wide range of texts, such as the Vedas, Upanishads, Puranas, and epic narratives like the Ramayana and Mahabharata. With the advent of the digital era, these scriptures have become globally accessible through digital libraries, online repositories, and other internet platforms. However, the complexities of Sanskrit syntax, particularly for Sanskrit poetry, can be daunting for those new to the language, creating a barrier to accessing these ancient texts. \\
Cross-Lingual Information Retrieval (CLIR) offers a promising solution to bridge this gap, enabling the retrieval of Sanskrit documents through English queries and thus broadening their accessibility to a diverse audience. Prior research has made considerable strides in Sanskrit text processing and retrieval. For instance, \cite{sahu2023building} developed and evaluated different indexing, stemming, and searching strategies specific to Sanskrit, including the proposal of novel stemmers which showed substantial improvements over traditional methods. Another work discussed the use of machine-readable Sanskrit texts and the challenges associated with traditional information retrieval systems, proposing strategies to improve retrieval efficiency. \\
Further enriching the landscape of Sanskrit retrieval systems, two notable systems have been developed, which utilize a sophisticated information retrieval architecture rather than conventional pattern matching tools or database systems. The Gaveṣikā system allows for the search of inflected forms of nominal or verbal stems and accommodates spelling variations by expanding the query stem to its inflected forms during search, although it does not cover phonetic transformations resulting from sandhi, impacting the system's recall (Srigowri and Karunakar, 2013). The SARIT corpus, on the other hand, implements a unique indexing strategy that supports document attributes, although its effectiveness is somewhat limited by the need for wildcard use in searches, impacting efficiency \cite{meyer2019sanskrit}. \\
Prior research on Sanskrit has predominantly focused on language processing tasks. \cite{krishna2021graph} developed an energy-based model framework for Word Segmentation, Morphological Parsing, and Dependency Parsing that requires minimal data, utilizing a search-based structured prediction approach. Additionally, \cite{sandhan2022translist} have explored the application of Transformers for Sanskrit Word Segmentation, achieving enhanced performance. There have also been significant strides in compound analysis; \cite{sandhan2022novel} introduced a compound type identification method that leverages contextual information and combines dependency parsing with morphological tagging. \cite{krishna2019poetry} devised a technique to transform Sanskrit poetry into prose format. Efforts towards named-entity recognition have also been noted, with \cite{sujoy2023pre} developing a pre-annotation method for a Named Entity Recognition (NER) dataset tailored for the Sanskrit Corpus. While these advancements have been crucial, relatively less focus has been placed on other significant tasks such as question answering and text summarization. Notably, Cross-Lingual Information Retrieval (CLIR), particularly involving Sanskrit, has remained largely unexplored, underlining the innovative nature of our Anveshana dataset in addressing this gap. \\
In response to the identified gaps and challenges in the field, and inspired by prior research in Sanskrit language processing, we embarked on a comprehensive benchmarking study to explore and evaluate current state-of-the-art models for Cross-Lingual Information Retrieval (CLIR) from English to Sanskrit. Our primary objective is to assess the effectiveness of these models in accurately retrieving Sanskrit documents based on English queries. To achieve this, we meticulously assembled a robust dataset, focusing on the Srimadbhagavatam, comprising 3,400 query-document pairs from 334 different documents. These documents were carefully curated to represent a wide spectrum of thematic content and complexity within the texts. Our dataset includes detailed preprocessing of Sanskrit documents to preserve their poetic structure while accommodating computational analysis, and minimal preprocessing of English queries to maintain their original intent. We employed a strategic approach to negative sampling in training our models, aiming to enhance their ability to discern relevant from non-relevant documents effectively. This meticulous preparation sets the stage for a nuanced exploration of CLIR, where the intersection of English queries and Sanskrit documents offers a rich landscape for examining linguistic and cultural transference. The evaluation of our models employs a comprehensive suite of metrics designed to measure their precision, recall, and overall accuracy in this unique retrieval context. Through rigorous testing and analysis, our study seeks to illuminate the capabilities and limitations of existing CLIR technologies applied to the rich but complex domain of Sanskrit texts. These insights are invaluable to researchers, technologists, and cultural scholars aiming to enhance the accessibility of these ancient texts through modern technological interventions. We plan to make both the dataset and the developed models publicly available to foster further research and development in the field. \\

We summarize our contributions as follows:
\begin{itemize}
        \item Developed \textit{Anveshana}, the first benchmark dataset tailored for Cross-Lingual Information Retrieval (CLIR) between English queries and Sanskrit documents, addressing the gap in resources for this ancient language.
        \item Implemented a series of advanced retrieval strategies that enhance both precision and recall, specifically designed to handle the complex syntactic and phonetic structures of Sanskrit.
        \item Evaluations using robust metrics such as NDCG, MAP, Recall, and Precision demonstrate that our approach outperforms existing methods, setting a new standard in CLIR for Sanskrit and potentially other ancient languages.
\end{itemize}

\section{Related Work}
Extensive research in Cross-Lingual Information Retrieval (CLIR) has paved the way for significant advancements in the field, utilizing robust neural language models and developing extensive multilingual resources. \cite{jiang2020cross} explored the application of BERT in CLIR, showcasing its effectiveness in aligning English queries with Lithuanian documents through a deep relevance matching model trained with weak supervision. Complementing these efforts, \cite{ogundepo2022africlirmatrix} introduced AfriCLIRMatrix, which broadens the linguistic diversity in CLIR resources by providing a dataset covering 15 African languages, aiming to spur further research in underrepresented languages. Additionally, \cite{sun2020clirmatrix} expanded available resources significantly by creating a vast collection of bilingual and multilingual datasets from Wikipedia, supporting end-to-end neural information retrieval across 19,182 language pairs.
The technical intricacies of CLIR have also been addressed through innovative approaches to document representation and retrieval strategies. \cite{yarmohammadi2019robust} developed robust document representations that combine N-best translations with bag-of-phrases outputs to enhance retrieval in low-resource settings, effectively handling errors from machine translation and automatic speech recognition. \cite{huang2021mixed} introduced the Mixed Attention Transformer (MAT), which leverages word-level external knowledge to bridge the translation gap in multilingual settings, significantly improving retrieval accuracy. Meanwhile, \cite{saleh2020document} analyzed the efficacy of document versus query translation approaches in medical CLIR, finding that query translation generally yields better retrieval results.
Further research explored adaptive and weakly supervised models to enhance CLIR’s effectiveness in resource-scarce environments. \cite{zhao2019weakly} proposed a weakly supervised model that utilizes translation corpora for training, focusing on attention mechanisms to identify relevant spans in foreign sentences, showing substantial improvements in retrieval accuracy. \cite{bi2020constraint} innovated in neural query translation by restricting the translation vocabulary to improve alignment with search indices, enhancing both translation and retrieval outcomes. \cite{boschee2019saral} presented SARAL, an end-to-end system for CLIR and summarization in low-resource languages, which demonstrated top performance in international evaluations.
Evaluative frameworks and augmented models also contributed to advancements in CLIR. \cite{sun2020clireval} developed CLIReval, a toolkit for evaluating machine translation within a CLIR framework, providing new metrics for assessing translation quality through its impact on retrieval effectiveness. Lastly, \cite{shi2023replug} introduced REPLUG, a retrieval-augmented language modeling framework that significantly enhances the capabilities of large-scale language models like GPT-3 by incorporating a tunable retrieval model, thereby improving language understanding and prediction accuracy.
These studies collectively provide a rich foundation for ongoing and future research in CLIR, showcasing a range of methodologies from resource development and innovative modeling to empirical evaluation, all of which significantly influence our project, Anveshana, in its goal to bridge the gap between English queries and Sanskrit documents.

\section{Dataset}
To effectively train and evaluate the necessary CLIR model, it was imperative to have Sanskrit documents paired with their English queries. We identified the Srimadbhagavatam as the only texts that provided the requisite data, with the Sanskrit documents being various chapters of the Srimadbhagavatam.

\subsection{Data Collection}
In the data collection phase of our CLIR research, we implemented web scraping techniques to harvest textual content from the website Vedabase\footnote{https://vedabase.io/en/library}. This digital platform hosts a variety of Sanskrit documents, including multiple chapters of the ancient text Srimadbhagavatam. For our study, we specifically focused on retrieving these documents to construct a robust dataset. To facilitate the development of query-document pairs essential for our cross-language retrieval tasks, taking inspiration from the work\cite{chen2017readingwikipediaansweropendomain}, we meticulously examined English translations of each document, and then manually crafted an average of 10 queries per document, resulting in a total of 3400 query-relevant document pairs across 334 documents. This curated dataset forms the backbone of our research, enabling us to rigorously test and refine our retrieval algorithms.

\subsection{Data Preparation}

\textbf{Preprocessing the Sanskrit Documents:} For the Sanskrit documents, we adhered to the preprocessing steps utilized in our previous work on Cross-Lingual Summarization. This involved the removal of numerical and punctuation characters not part of the Devanagari script. Special attention was given to poetic structures, where sentence demarcations were marked by sequences like ”||1.1.3||” in the texts. Using Python's regex capabilities, these markers were substituted with a single "|", facilitating sentence splitting. Despite the conventional preference for converting poetic Sanskrit to prose for computational ease, our dataset maintained the original poetic format to preserve linguistic nuances. \\
\textbf{Handling English Queries:} The English queries presented a different challenge due to their language and format. Given their brevity and the lesser complexity in terms of structural tokens compared to the Sanskrit documents, the English queries required no extensive preprocessing. This decision was made to retain the queries' original intent and complexity, ensuring a more authentic cross-lingual retrieval task. \\

\begin{figure*}[ht]
    \centering
    \begin{subfigure}{0.32\linewidth}
        \includegraphics[width=\linewidth,height=6cm]{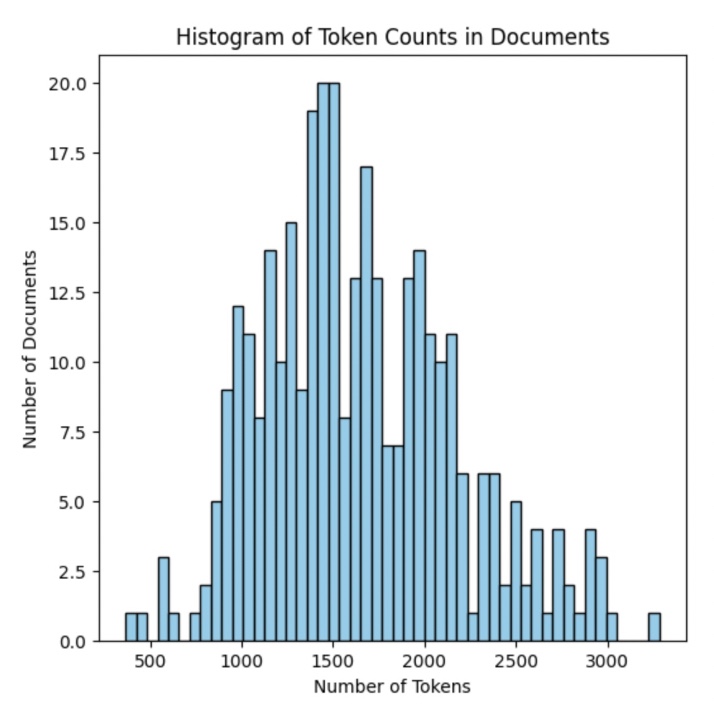} % Replace with your image file
        \caption{Token count distribution in Sanskrit documents}
        \label{fig:DocTokenCounts}
    \end{subfigure}
    \begin{subfigure}{0.32\linewidth}
        \includegraphics[width=\linewidth,height=6cm]{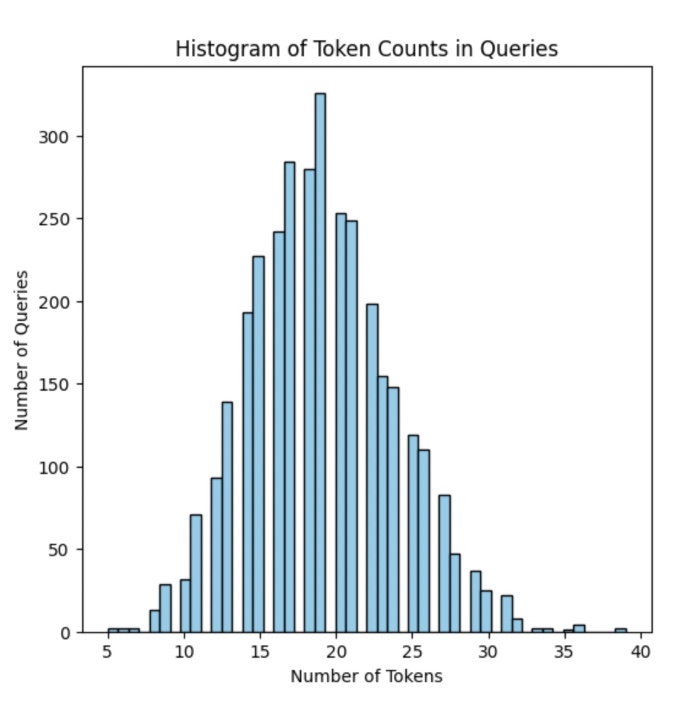} % Replace with your image file
        \caption{Token count distribution in English queries}
        \label{fig:QueryTokenCounts}
    \end{subfigure}
    \begin{subfigure}{0.32\linewidth}
        \includegraphics[width=\linewidth,height=6cm]{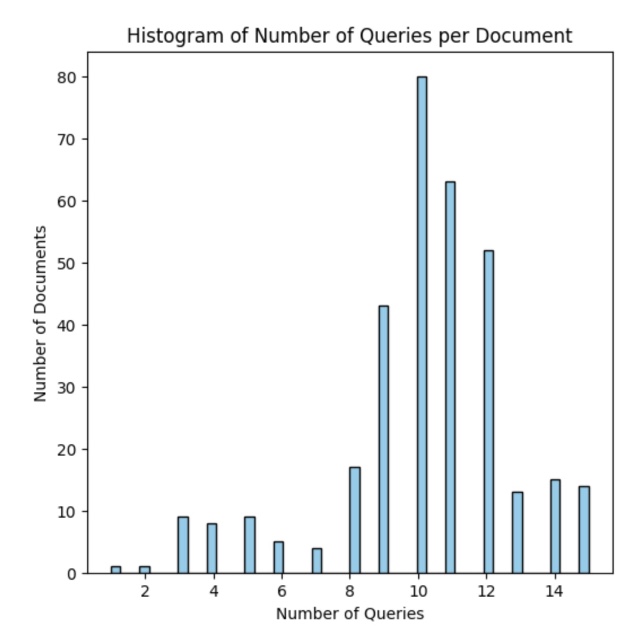} % Replace with your image file
        \caption{Distribution of query counts per document}
        \label{fig:QueryCountsPerDoc}
    \end{subfigure}
\end{figure*}

\textbf{Dataset Composition and Negative Sampling:} Our dataset comprised 3400 pairs of English queries and relevant Sanskrit documents, sourced from a collection of 334 distinct Sanskrit texts. We split the data in 90:10 ratio to create, train, and test dataset of size 3060 and 340 query-document pairs, respectively. To enhance the model's discrimination capabilities, we employed a negative sampling strategy on the train dataset. This approach involved generating non-relevant query-document pairs, training the model to distinguish between relevant and non-relevant matches. The training employed a 2:1 ratio of negative to positive samples, aiming to robustly tune the model's relevance detection in a cross-lingual context. The new  comprehensive training dataset, that consists of the original query-document pairs along with negative samples of non-matching query-document pair, was split using 90:10 ratio, thus creating a train and validation dataset. All the splits: train, validation and test, the total number of documents is same as pre-split i.e. 334.

\subsection{Data Analysis}

In our Cross-Language Information Retrieval (CLIR) project, we've conducted an analysis of token distributions within both the queries and documents to facilitate a deeper understanding of the dataset dynamics. The corpus consists of 334 Sanskrit documents with a token range stretching from a minimum of 365 to a maximum of 3286, averaging at 1645.41 tokens per document. The distribution of tokens among documents exhibits a right-skewed pattern, signifying a concentration of documents with token counts below the mean. In contrast, the English queries display a distinct pattern where the majority are succinct, with a sharp peak in lower token counts—ranging from 5 to 39 tokens and averaging 19.05 tokens per query. This stark difference in token distribution between the verbose Sanskrit documents and the concise English queries presents a unique challenge for the retrieval system, underscoring the necessity for effective translation and token normalization strategies in the CLIR framework. 

Figure \ref{fig:DocTokenCounts} shows a right-skewed token distribution in Sanskrit documents, indicating a prevalence of shorter documents in the corpus. Figure \ref{fig:QueryTokenCounts} depicts an almost normal distribution of token counts in English queries, suggesting a consistency in query length. Lastly, Figure \ref{fig:QueryCountsPerDoc} illustrates a left-skewed distribution of query counts per document, revealing that most Sanskrit documents are associated with a larger number of queries.

% Here, insert the figures for CLIR Dataset
\begin{table}[h]
    \centering
    \resizebox{!}{0.90cm}{
    \begin{tabular}{|c|c|c|}
        \hline
        \textbf{Statistics} & \textbf{Documents} & \textbf{Queries} \\
        \hline
        Count & 334 & 3400 \\
        \hline
        Maximum Token count & 3286 & 39 \\
        \hline
        Minimum Token count & 365 & 5 \\
        \hline
        Average Token count & 1645.41 & 19.05 \\
        \hline 
    \end{tabular}
    }
    \caption{Statistics of Documents and Queries for CLIR Dataset}
    \label{tab:statistics}
\end{table}

\subsection{Dataset Example}
In the provided data sample Table~\ref{table:clir_sample1}, the English query asks about the nine different ways of rendering devotional service according to the Srimad Bhagavatam, a central text in devotional Hinduism. The corresponding Sanskrit document, cited in the table, begins with a discourse by King Rahugana, addressing fundamental philosophical inquiries that are indirectly related to the query. This excerpt from the Srimad Bhagavatam elaborates on profound metaphysical concepts and the nature of reality as perceived through the lens of Vedanta philosophy, encapsulating the essence of devotional service through a philosophical dialogue. This juxtaposition of a direct query with a philosophically rich text exemplifies the dataset's potential in providing comprehensive answers that not only address the query’s factual demands but also offer deeper insights into the broader philosophical and theological contexts. This approach enhances the educational and research utility of the "Anveshana" dataset, making it a valuable resource for those seeking to explore and understand Sanskrit scriptures through English queries.

\begin{table*}
\begin{center}
\centering
\resizebox{!}{5.35cm}{
\begin{tabular}{|p{5cm}|p{4cm}|} 
\hline
\textbf{Query} & \textbf{Document} \\ [0.5ex] 
\hline
What are the nine different ways of rendering devotional service, according to Srimad Bhagavatam? & \begin{minipage}{.3\textwidth}
       \vspace{2pt}
      \includegraphics[width=4cm, height=6.0cm]{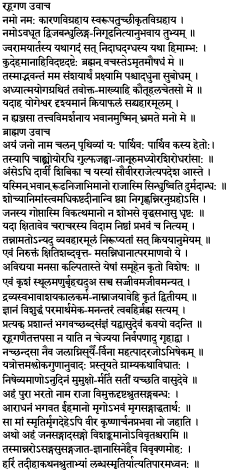}
    \end{minipage} \\
\hline
\end{tabular}
}
\caption{\label{table:clir_sample1}
This table presents a sample query from the Anveshana dataset paired with its corresponding Sanskrit document, illustrating the dataset's application in enhancing cross-lingual information retrieval capabilities for English to Sanskrit retrieval.}
\end{center}
\end{table*}

\section{Methodology}
In our methodology for the pioneering English-Sanskrit CLIR dataset, we strategically bypass the intricate challenges of direct translation by leveraging the Google Translate API for both query and document translation tasks, acknowledging the current limitations in handling the complexity of the Sanskrit language, a notable low-resource linguistic domain. Within this framework, we employ a multifaceted approach to tackle the broader challenges of CLIR, which include navigating linguistic nuances, differing grammatical structures, and variations in the level of detail between languages. Our methodology encompasses three primary strategies:

\textbf{Query Translation (QT):} We adopt the QT approach by utilizing the Google Translate API to convert English queries into Sanskrit, facilitating monolingual retrieval within the Sanskrit document corpus. This step is crucial in overcoming the initial language barrier and setting the stage for effective information retrieval.

\textbf{Document Translation (DT):} In parallel, the DT approach is employed where Sanskrit documents are translated into English, again leveraging the Google Translate API. This enables the evaluation and relevance assessment of retrieved documents in the query's native language, streamlining the information retrieval process.

\textbf{Direct-Retrieve(DR):} Additionally, we explore the DR method, wherein Sanskrit documents are directly retrieved in response to English queries through a shared embedding space. This approach seeks to bypass the intricacies of translation, relying on the semantic alignment of languages within a multidimensional vector space.

\begin{figure*}[ht]
    \centering
    % First row, first image
        \includegraphics[scale = 0.50]{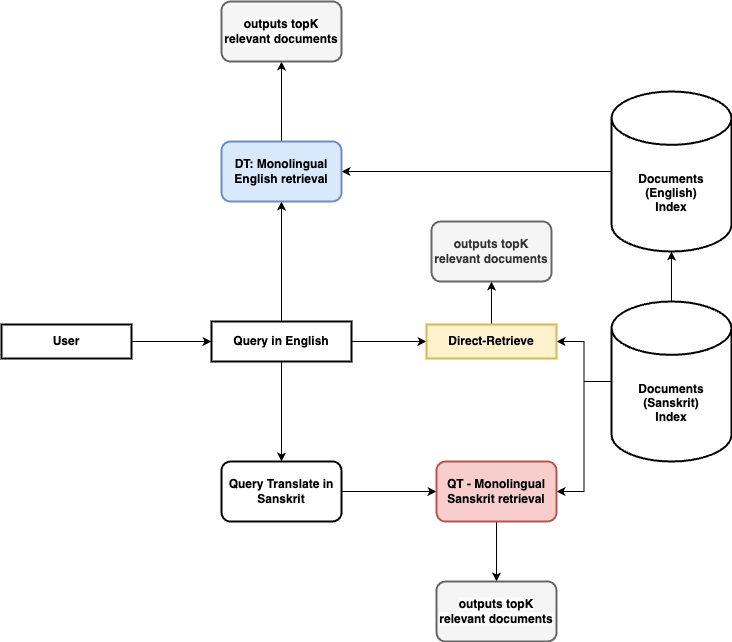}
        \caption{A flowchart to depict the different frameworks
        in CLIR: Query Translation (QT), Document Translation (DT), and Direct-Retrieve (DR)}
        \label{fig:CLIR-flowchart}
    \vspace{1cm} % Ensures that the following figure is set to 
\end{figure*}

In Figure \ref{fig:CLIR-flowchart}, we present a flowchart depicting the different frameworks in Cross-Lingual Information Retrieval (CLIR): Query Translation (QT), Document Translation (DT), and Direct-Retrieve (DR). In our experiments, we evaluate the efficacy of several models across three distinct frameworks for Cross-Lingual Information Retrieval (CLIR):\\

\textbf{Query Translation (QT):}
\begin{itemize}
    \item BM25 \cite{robertson2009probabilistic}: Employed for its effectiveness in monolingual retrieval after translating English queries into Sanskrit.
    \item Xlm-roberta-base \cite{conneau2019unsupervised}: Utilized to optimize retrieval performance specifically for Sanskrit documents.
\end{itemize}

\textbf{Document Translation (DT):}
\begin{itemize}
    \item BM25 \cite{robertson2009probabilistic}: Applied for its robustness in retrieving documents after translating Sanskrit texts into English.
    \item contrieverCAT \cite{izacard2021unsupervised}: Optimizes retrieval by improving the semantic understanding between queries and documents.
    \item contrieverDOT \cite{izacard2021unsupervised}: Further refines retrieval accuracy through advanced embedding techniques.
    \item colbert \cite{khattab2020colbert}: Enhances document retrieval through fine-tuned contextual embeddings.
    \item GPT2 \cite{radford2019language}: Integrated to process complex query-document interactions and improve retrieval outcomes.
    \item REPLUG LSR \cite{shi2023replug}: Leverages a retrieval-augmented language model to refine and enhance document selection based on query relevance.
\end{itemize}

\textbf{Direct Retrieve (DR):}
\begin{itemize}
    \item Baseline neural CLIR model \cite{sun2020clirmatrix}: Utilizes a multidimensional retrieval approach to process queries and documents within shared embedding spaces, facilitating direct retrieval without the need for translation.
    \item Xlm-roberta-base \cite{conneau2019unsupervised}: Employs advanced multilingual capabilities to directly match English queries with Sanskrit documents within a unified embedding space.
    \item Intfloat/multilingual-e5-base \cite{wang2024multilingual}: Engineered for cross-lingual compatibility and semantic understanding, this model facilitates the direct retrieval of documents by interpreting and processing queries in multiple languages.
    \item GPT2 \cite{radford2019language}: Supports complex semantic processing to directly retrieve documents based on query content.
    \end{itemize}
\textbf{Zero-shot Models:}
\begin{itemize}
    \item Colbert \cite{khattab2020colbert}: Provides robust zero-shot capabilities for document retrieval by generating context-rich embeddings without specific fine-tuning on CLIR tasks.
    \item Xlm-roberta-base \cite{conneau2019unsupervised}: Offers extensive multilingual support, facilitating direct and zero-shot retrieval across different language pairs.
    \item Contriever \cite{izacard2021unsupervised}: Applied for retrieving English documents directly from English queries without translation, demonstrating effective zero-shot retrieval capabilities.
    \item Intfloat/multilingual-e5-base \cite{wang2024multilingual}: Optimized for zero-shot multilingual document retrieval, this model can process and understand queries in various languages, enabling efficient and contextually aware retrieval without the need for task-specific tuning.

\end{itemize}

\subsection{Query Translation}
\begin{itemize}
    \item \textbf{BM25} \cite{robertson2009probabilistic}:
    BM25 is an advanced retrieval function that ranks documents based on how often the query terms occur within them, considering both the document length and the overall term frequency across the corpus. It utilizes term frequency (TF), which is the number of times a term appears in a document, and inverse document frequency (IDF), which measures the rarity of a term among all documents. BM25 adjusts for document length to prevent long documents from being overweighted merely because they contain more words. This balance makes BM25 particularly useful in situations requiring precise term relevance, such as retrieving documents in a specific language like Sanskrit, where both queries and documents are in the same language, thus ensuring streamlined retrieval focusing on term relevance and document relevancy without language translation barriers.

    \item \textbf{Xlm-roberta-base} \cite{conneau2019unsupervised}:
    The xlm-roberta-base model, fine-tuned for the Sanskrit language, adeptly handles its unique grammatical and syntactic features to enhance retrieval accuracy. It generates embeddings for queries and documents and calculates similarity by taking the dot product of these embeddings, scaled by \(\frac{1}{\sqrt{\text{length of embeddings}}}\), to normalize the effect of dimensionality. A sigmoid function then transforms this value into a relevance score between 0 and 1, indicating the likelihood of relevance. This model is trained in a supervised manner using a dataset with columns for query, document, and a binary label (1 for relevant, 0 for not relevant) using Binary Cross-Entropy (BCE) as the loss function. Alternatively, the model can also be trained by concatenating the query and document with a [SEP] separator, treating the task as a binary classification problem to directly determine the relevance of the document to the query.
\end{itemize}

\subsection{Document Translation}
\begin{itemize}
    \item \textbf{mjwong/contriever-mnli fine-tuned (CONCAT and DOT)} \cite{izacard2021unsupervised}: We fine-tune contriever by concatenating the query and document with a [SEP] separator, treating the retrieval task as a binary classification problem that directly determines the relevance of the document to the query. This model utilizes a bi-encoder architecture that encodes queries and documents independently, allowing relevance scores to be computed via the dot product of their embeddings. It incorporates contrastive learning in an unsupervised setting to improve its discriminative capabilities; this involves optimizing a contrastive loss function that differentiates between relevant (positive) and irrelevant (negative) document pairs by maximizing the distance between them in the embedding space. Positive pairs are generated from closely related document segments, while negative pairs are broadly sampled from the dataset, thus refining the model's accuracy in identifying document relevance across various retrieval tasks. Similarly to the xlm-roberta-base, we fine-tune the CONTRIEVER model by generating embeddings from Sanskrit queries and documents. We scale the dot product of these embeddings by the inverse square root of the embedding length and then apply a sigmoid function to derive a relevance score between 0 and 1. We train this model using a dataset labeled with binary relevance, employing Binary Cross-Entropy (BCE) as the loss function to actively optimize the model’s ability to distinguish between relevant and irrelevant documents.

    \item \textbf{ColBERT} \cite{khattab2020colbert}: ColBERT, or Columnar BERT, is specifically designed for document retrieval by enhancing the traditional BERT architecture to include a late interaction mechanism, making it highly effective in monolingual settings. This model processes each token of the input query and document independently to produce separate embeddings, allowing for a detailed comparison via a dot product between each token of the query and the document. Such comparisons are aggregated to form a comprehensive relevance score. To fine-tune ColBERT on our dataset, we employ the dot product of token embeddings to quantify semantic similarity at a granular level, optimizing overall document relevance. During training, we minimize the cross-entropy loss between predicted relevance scores and actual labels, refining the attention mechanism to accurately emphasize the most significant tokens, thus enhancing the model’s retrieval accuracy in monolingual environments.

    \item \textbf{GPT-2} \cite{radford2019language}: GPT-2, developed by OpenAI, is a transformer-based language model pre-trained on a broad corpus of internet text using an unsupervised learning method focused on predicting the next word in a sentence. Leveraging its capacity to understand complex language patterns, we adapted GPT-2 for document retrieval by fine-tuning it on our dataset, which includes query-document pairs labeled for relevance. During fine-tuning, GPT-2 processes the concatenation of each query and document, separated by a [SEP] delimiter, to maintain context distinction. It then applies logistic regression over its final layer outputs to generate a relevance score for each pair, optimizing the binary cross-entropy loss between predicted scores and actual labels. This approach harnesses GPT-2’s deep textual understanding to effectively identify relevant documents in response to queries, enhancing retrieval accuracy within our dataset.

    \item \textbf{REPLUG LSR} \cite{shi2023replug}: REPLUG LSR adapts the dense retriever model by utilizing the language model (LM) to provide supervision, aiming to retrieve documents that result in lower perplexity scores, effectively guiding the retriever towards more relevant content. This training involves four main steps: retrieving documents with the highest similarity scores, scoring these documents using the LM based on how well they improve LM perplexity, updating the retrieval model parameters by minimizing the Kullback-Leibler (KL) divergence between the retrieval likelihood and the LM’s score distribution, and asynchronously updating the datastore index to keep the document embeddings current. In our implementation for the Document Translation (DT) approach, we utilized the contriever model as a trainable retriever. We labeled our dataset using Mistral-7B to generate ground truth continuations (y) and configured the retrieval process to mirror REPLUG LSR's method. For each query, after retrieving the top 20 documents based on their similarity, we concatenated each query with the retrieved documents separately and passed them through GPT-2. We then calculated the loss between the LM-generated text and the Mistral-7B-generated ground truth. This loss was scaled and negated to serve as a basis for calculating the LM likelihood via a softmax function, reflecting that lower losses (closer generated and ground truth texts) correspond to higher relevance. The LM likelihood for each document was then computed as the softmax of the negative scaled loss, which prioritizes documents that are semantically closer to the ground truth. Finally, we minimized the KL divergence between the retrieval likelihood and LM likelihood to fine-tune the model, ensuring that our retrieval system effectively identifies and prioritizes documents most relevant to the query.
\end{itemize}

\subsection{Direct-Retrieve}
\begin{itemize}
    \item In our Direct Retrieve (DR) framework, we employed the mDPR (Multi-Dimensional Passage Retrieval) model, which embeds queries and documents into a shared high-dimensional space. This approach is particularly advantageous in multilingual settings as it bypasses the need for language translation, focusing instead on capturing semantic similarities directly. To enhance this capability, we fine-tuned the xlm-roberta-base \cite{conneau2019unsupervised} model, known for its proficiency in handling diverse languages. This fine-tuning process aimed to improve the model’s ability to discern and match queries and documents based on deep linguistic and semantic analysis, effectively boosting the retrieval performance in our multilingual dataset.
    \item The intfloat/multilingual-e5-base \cite{wang2024multilingual} is part of a trio of multilingual embedding models designed to balance the trade-offs between inference efficiency and the quality of embeddings. This model, particularly the 'base' variant, has been engineered using a multi-stage training pipeline. Initially, it undergoes weakly-supervised contrastive pre-training on approximately 1 billion text pairs, sourced from a wide array of datasets such as Wikipedia, mC4, Multilingual CC News, and others listed in the study. This stage employs large batch sizes and leverages the standard InfoNCE contrastive loss with in-batch negatives, aligning it closely with procedures used for English E5 models. Following this, the model is fine-tuned on a combination of high-quality labeled datasets, incorporating both in-batch and mined hard negatives, as well as knowledge distillation from a cross-encoder model. This fine-tuning is aimed at significantly enhancing the model’s capability to discern deep linguistic and semantic nuances across languages, which is critical for applications like multilingual document retrieval where the model has to deal with complex queries in English and document content in Sanskrit. The detailed performance metrics post-fine-tuning, as illustrated in your study, demonstrate the efficacy of this model in improving retrieval performance within multilingual settings, marking it as a robust solution for enhancing the precision and effectiveness of cross-lingual information retrieval systems.
    \item Additionally, we integrated GPT-2 \cite{radford2019language} into our retrieval process using several approaches: First, by concatenating English queries with Sanskrit documents and fine-tuning xlm-roberta-base as a binary classification task, aiming to directly determine the relevance of document-query pairs. Second, we extracted embeddings from the final layer of the model, calculated their dot product, scaled these values to a 0-1 range, and treated the resulting scores as binary classification labels. Lastly, to further refine our model, we incorporated hard negative samples generated by BM25 \cite{robertson2009probabilistic} into our training set. This approach, using both xlm-roberta-base and GPT-2, helped fine-tune the models to better distinguish between closely related but non-relevant documents, thereby enhancing the precision of our retrieval system.
\end{itemize}

In our zero-shot retrieval setup, we utilized a suite of models to evaluate their ability to effectively retrieve documents without further fine-tuning on our specific dataset. The colbert \cite{khattab2020colbert}, which leverages pre-trained embeddings to assess document relevance, providing robust zero-shot capabilities especially useful in scenarios with limited labeled data. The xlm-roberta-base \cite{conneau2019unsupervised} model, known for its multilingual capabilities, was employed to handle documents and queries across various languages directly, serving as a versatile tool for initial retrieval phases. Additionally, the contriever \cite{izacard2021unsupervised} was used for its inherent understanding of English to perform zero-shot retrieval of English documents.
For the retrieval process, we first embedded documents and queries using these models, then calculated the dot product of these embeddings to evaluate semantic similarities. The results were sorted to identify the top-K documents most relevant to each query, effectively utilizing zero-shot capabilities to predict relevance based on pre-learned language representations. To facilitate efficient retrieval of the top-K results, we employed a Faiss index \cite{douze2024faiss}, a library for efficient similarity search and clustering of dense vectors, which enhances the scalability and speed of our retrieval process, especially in large-scale datasets. This combination of zero-shot models and efficient indexing technology provided a comprehensive approach to evaluating and improving document retrieval without task-specific training.

\section{Experiments}
In this section, we explain the evaluation metrics employed for the purposes of bench-marking and a detailed explanation of our experimental setup.
\subsection{Evaluation Metrics}
To effectively evaluate the performance of our Cross-Lingual Information Retrieval (CLIR) model, we utilize a suite of established metrics that comprehensively assess both the relevance and ranking quality of the search results provided by our system. The key metrics employed include Recall, Precision, Normalized Discounted Cumulative Gain (NDCG), and Mean Average Precision (MAP) at various cut-off points: k = 1, 3, 5, and 10. Recall \cite{arora2016evaluation} measures the proportion of relevant documents that are successfully retrieved by the model out of all relevant documents available in the dataset. It provides an indication of the model’s ability to retrieve all pertinent information without missing significant documents. Precision \cite{arora2016evaluation}, on the other hand, evaluates the fraction of retrieved documents that are relevant. This metric is crucial for understanding how effectively the model avoids fetching irrelevant documents, thereby ensuring the precision of the retrieval process. NDCG \cite{wang2013theoretical} is particularly vital in scenarios where the relevance of retrieved documents is not binary but graded. This metric assesses the quality of the ranking by rewarding highly relevant documents appearing earlier in the search results list more than those appearing later. NDCG is normalized against the ideal possible gain, making it a robust indicator of ranking effectiveness across different query sets. MAP \cite{revaud2019learning} measures the average precision at various cut-off points in the ranking process, specifically at k = 1, 3, 5, and 10 in our evaluation. MAP provides a comprehensive view of precision at multiple levels of retrieval depth, accommodating diverse user interactions ranging from shallow to deep dives into the search results. Utilizing these metrics together allows for a robust and nuanced understanding of a CLIR system’s performance.

\subsection{Experimental Setup}
In our pioneering study on English-Sanskrit Cross-Lingual Information Retrieval (CLIR), we implemented a rigorous experimental setup to explore and validate the effectiveness of our methodologies across three frameworks: Query Translation (QT), Document Translation (DT), and Direct-Retrieve (DR). \\

In the QT framework, we utilize the Google Translate API to convert English queries into Sanskrit, setting the stage for monolingual retrieval processes. Subsequently, we process these translated queries using the BM25 model, which ranks documents based on the frequency of query terms while adjusting for document length and term frequency. Alongside this, we deploy a specialized monolingual retrieval model tailored for the Sanskrit language. This model generates embeddings from the translated queries and computes relevance scores by scaling the dot product of embeddings through the inverse square root of their lengths, followed by a sigmoid transformation to finalize scores between 0 and 1. We train this model on a dataset labeled with binary relevance indicators, optimizing performance using Binary Cross-Entropy as the loss function to accurately match queries to relevant documents.

For the DT framework, we translate entire Sanskrit documents into English via Google Translate, facilitating the application of English-specific monolingual retrieval techniques. Initially, the BM25 algorithm ranks these translated documents based on textual relevance. To further enhance retrieval, we employ several models: Best Monolingual Retrieval for English processes concatenated text, optimizing through binary classification with Cross-Entropy loss. Both the \texttt{mjwong/contriever-mnli} and \texttt{contriever} models, fine-tuned using CONCAT and DOT methods, independently encode queries and documents to refine their linguistic matching capabilities. Additionally, \texttt{MonolingualColBERT\_eng} and \texttt{GPT2} generate deep contextual embeddings, with \texttt{GPT2} also handling concatenated text as binary classification tasks. \texttt{REPLUG LSR} utilizes a language model for supervision, training to minimize the Kullback-Leibler divergence between model predictions and true labels, thereby increasing retrieval precision.

In the DR framework, we leverage the mDPR model to embed queries and documents into a unified high-dimensional space, effectively bypassing language barriers and focusing on semantic similarities. To enhance this model's functionality, we have also integrated the intfloat/multilingual-e5-base model, which has been specifically fine-tuned to improve handling of multilingual content. This process involves projecting embeddings into a common space and utilizing cosine similarity measures for semantic matching. The intfloat/multilingual-e5-base model, known for its robust performance in multilingual settings, undergoes a sophisticated fine-tuning regimen. This includes training on a rich mix of multilingual text pairs and further refined by supervised fine-tuning using high-quality labeled datasets to improve its semantic understanding across languages. This dual-model approach ensures that our embeddings capture the nuanced meanings of English queries and Sanskrit documents, significantly enhancing the precision and efficacy of our retrieval system. Additionally, we incorporate processes from GPT-2 to handle concatenated English queries and Sanskrit documents, fine-tuning them within a binary classification framework using Binary Cross-Entropy loss, thereby further boosting the system's ability to distinguish between relevant and non-relevant documents in our multilingual retrieval tasks. This comprehensive embedding and fine-tuning strategy, combining mDPR, xlm-roberta-base, and intfloat/multilingual-e5-base, forms the core of our experimental section, demonstrating a significant leap in the performance of our CLIR system.

\section{Results}
In this section, we present the results from our comprehensive evaluation of various models and frameworks developed for cross-lingual information retrieval (CLIR) on the test dataset. Our systematic approach involved rigorous testing across different settings to discern the performance capabilities of each model within the frameworks of Query Translation (QT), Document Translation (DT), Direct Retrieve (DR), and Zero-shot applications.

In the zero-shot framework of the Cross-Lingual Information Retrieval (CLIR) project, the \textit{monolingual-Contriever\_eng} model outperformed other models with superior results across multiple metrics, achieving NDCG scores of 25.09\%, 27.88\%, and 30.48\% at cutoffs of $k = 3$, $k = 5$, and $k = 10$ respectively, along with corresponding Recall scores of 30.39\%, 37.23\%, and 45.30\%. The \textit{monolingual-ColBERT\_eng\_mean
\_last\_hidden\_state} and \textit{monolingual-ColBERT\_eng\_CLS\_embedding} models also showed commendable performance, with the former model recording NDCG of 17.18\% at $k = 3$, progressively increasing to 21.77\% at $k = 10$. These results underscore the robustness of pretrained models in zero-shot retrieval tasks, particularly highlighting the effectiveness of the \textit{monolingual-Contriever\_eng} in handling document retrieval without any fine-tuning across varying evaluation metrics.

In the QT (Query Translation) framework, BM25 demonstrated a uniform performance across all metrics at approximately 2.95\%, while the \textit{monolingual-xlm-roberta-base\_sanskrit} trailed with about 0.14\%. This underscores the challenges in optimizing monolingual retrieval in Sanskrit without translation techniques.

In the DT (Document Translation) framework, BM25 demonstrated exceptionally high performance, with notable scores of 56.04\% in NDCG@3, 59.76\% in NDCG@5, and 62.46\% in NDCG@10, effectively showcasing its strong adaptability and effectiveness in handling translated content for retrieval purposes. The \textit{ColBERT} fine-tuned with the \textit{DOT} method also showed significant results, achieving 33.12\% in NDCG@3, 37.52\% in NDCG@5, and 40.78\% in NDCG@10, while the \textit{contriever} fine-tuned with \textit{DOT} method yielded 34.42\% in NDCG@3, 38.29\% in NDCG@5, and 41.70\% in NDCG@10. Additionally, the \textit{REPLUG LSR}, incorporating \textit{Mistral-7B-Instruct-v0.2} and \textit{GPT-sup}, demonstrated its efficacy with 21.22\% in NDCG@3, 23.10\% in NDCG@5, and 26.26\% in NDCG@10, underscoring the effectiveness of advanced retrieval techniques integrated with language model supervision.

The DR (Direct Retrieve) framework demonstrated varied results, with the \textit{mDPR-BM35-HN1} showing modest performance, achieving 3.87\% in NDCG@3, 4.22\% in NDCG@5, and 4.74\% in NDCG@10, which reflects the challenges of direct retrieval without translation. The \textit{GPT2} model also showed limited effectiveness with 1.88\% in NDCG@3, 2.09\% in NDCG@5, and 2.54\% in NDCG@10. In contrast, the application of fine-tuned \textit{xlm-roberta-base} models in different configurations (\textit{DOT} and \textit{CONCAT}) underscored the nuanced capabilities of neural embeddings in enhancing direct multilingual retrieval, with the \textit{DOT} configuration achieving 3.24\% in NDCG@3, 3.98\% in NDCG@5, and 4.26\% in NDCG@10, and the \textit{CONCAT} configuration yielding 3.01\% in NDCG@3, 3.19\% in NDCG@5, and 3.53\% in NDCG@10. Additionally, the \textit{intfloat/multilingual-e5-base} showed promising improvements with 5.92\% across all NDCG measures at k=3, 8.86\% at k=5, and 10.74\% at k=10, illustrating the potential of integrating advanced model architectures in direct retrieval settings.

Continuing from the evaluation of retrieval performance at $k=1$, as detailed in our previous discussions, we observed distinct patterns of performance enhancement as $k$ values increased. Notably, as we expanded the evaluation to $k=3$, $k=5$, and $k=10$, the performance metrics generally showed upward trends across most models, demonstrating the benefit of considering a broader set of retrieved documents. When comparing metrics across these $k$ values, we see substantial improvements, particularly in the Zero-shot and DT frameworks. For example, the \textit{monolingual-Contriever\_eng} model increased from 18.33\% at $k=1$ to 25.09\% at $k=3$, further to 27.88\% at $k=5$, and a notable 30.48\% in NDCG at $k=10$. This increase reflects the model's robustness and its ability to effectively rank highly relevant documents even in a broader retrieval context. Similarly, the BM25 model in the DT framework showcased a significant rise from 40.64\% at $k=1$ to 56.04\% at $k=3$, 59.76\% at $k=5$, and 62.46\% at $k=10$ in NDCG, illustrating the effectiveness of traditional retrieval methods when adapted to cross-lingual settings and evaluated over a larger set of top retrieved documents.
Comparing across different metrics such as NDCG, MAP, Recall, and Precision for any given k, each metric provides insights into different aspects of retrieval quality. For instance, NDCG offers a view of the overall ranking quality, incorporating the position of relevant documents, whereas Recall measures the model's ability to retrieve all relevant documents, and Precision focuses on the accuracy of the retrieval in the top-k results. MAP, providing an average precision across queries, offers a cumulative measure of performance across the dataset.
The consistent improvement in these metrics as k increases underscores the models' ability to capture relevant documents even if they are not ranked at the very top. This observation is crucial for applications where the retrieval system can present a larger set of results for user refinement or automated processing. Furthermore, the differentiation in performance across models at varying k values and metrics highlights the nuanced effectiveness of each model and framework, guiding us in optimizing and selecting appropriate models for specific CLIR applications.

\begin{table*}[h]
\centering
\resizebox{\textwidth}{!}{%
\begin{tabular}{|l|l|c|c|c|c|}
\hline
\textbf{Framework} & \textbf{Model} & \textbf{\{NDCG, MAP, Recall, Precision\}@1} \\ \hline
Zero-shot & monolingual-ColBERT\_eng\_mean\_last\_hidden\_state & 12.68  \\ \cline{2-3} 
 & monolingual-ColBERT\_eng\_CLS\_embedding & 11.40 \\ \cline{2-3} 
 & xlm-roberta-base & 1.50 \\ \cline{2-3}
 & intfloat/multilingual-e5-base & 3.67 \\ \cline{2-3}
 & monolingual-Contriever\_eng & 18.33 \\ \hline
QT & BM25 & 2.95 \\ \cline{2-3} 
 & monolingual-xlm-roberta-base\_sanskrit & 0.14 \\ \hline
DT & BM25 & \textbf{40.64} \\ \cline{2-3} 
 & ColBERT-fine-tuned\_DOT & 21.22 \\ \cline{2-3} 
 & contriever - fine-tuned\_DOT & 23.50 \\ \cline{2-3} 
 & GPT2 & 1.43 \\ \cline{2-3} 
 & REPLUG LSR: Mistral-7B-Instruct-v0.2 and GPT2-sup & 14.87 \\ \hline
DR & Fine-tuned-xlm-roberta-base\_DOT & 2.16 \\ \cline{2-3} 
 & Fine-tuned-xlm-roberta-base\_CONCAT & 2.48 \\ \cline{2-3} 
 & intfloat/multilingual-e5-base & 5.92 \\ \cline{2-3} 
 & mDPR-BM35-HN1 & 2.94 \\ \cline{2-3} 
 & GPT2 & 1.52 \\ \hline
\end{tabular}%
}

\caption{Retrieval performance metrics for $k = 1$. For $k = 1$, NDCG, MAP, Precision,Recall will yield same value.  The table presents NDCG, MAP, Recall, and Precision, for $k = 1$, across various retrieval models within the Query Translation (QT), Document Translation (DT), Direct Retrieve (DR), and Zero-shot frameworks. This comprehensive summary illustrates the effectiveness of each model in handling different aspects of cross-lingual information retrieval.}
\label{table:retrieval-metrics-k1}
\end{table*}

\begin{table*}[h]
\centering
\resizebox{\textwidth}{!}{%
\begin{tabular}{|l|l|c|c|c|c|}
\hline
\textbf{Framework} & \textbf{Model} & \textbf{NDCG@3} & \textbf{MAP@3} & \textbf{Recall@3} & \textbf{Precision@3} \\ \hline
Zero-shot & monolingual-ColBERT\_eng\_mean\_last\_hidden\_state & 17.18 & 16.07 & 20.37 & 6.79 \\ \cline{2-6} 
 & monolingual-ColBERT\_eng\_CLS\_embedding & 15.80 & 9.15 & 19.28 & 6.43 \\ \cline{2-6} 
 & xlm-roberta-base & 2.02 & 1.88 & 2.41 & 1.37 \\ \cline{2-6} 
 & intfloat/multilingual-e5-base & 3.67 & 3.67 & 3.67 & 1.22 \\ \cline{2-6}
 & monolingual-Contriever\_eng & 25.09 & 14.99 & 30.39 & 10.13 \\ \hline
QT & BM25 & 5.18 & 4.09 & 5.63 & 1.88 \\ \cline{2-6} 
 & monolingual-xlm-roberta-base\_sanskrit & 0.56 & 0.36 & 0.85 & 0.28 \\ \hline
DT & BM25 & \textbf{56.04} & \textbf{48.45} & \textbf{58.24} & \textbf{19.41} \\ \cline{2-6} 
 & ColBERT-fine-tuned\_DOT & 33.12 & 30.61 & 38.46 & 12.21 \\ \cline{2-6} 
 & contriever - fine-tuned\_DOT & 34.42 & 31.79 & 42.02 & 14.01 \\ \cline{2-6} 
 & GPT2 & 1.61 & 1.57 & 1.71 & 1.24 \\ \cline{2-6} 
 & REPLUG LSR: Mistral-7B-Instruct-v0.2 and GPT2-sup & 21.22 & 19.56 & 26.07 & 8.69 \\ \hline
DR & Fine-tuned xlm-roberta-base\_DOT & 3.24 & 2.82 & 3.10 & 1.86 \\ \cline{2-6} 
 & Fine-tuned xlm-roberta-base\_CONCAT & 3.01 & 2.42 & 2.22 & 1.52 \\ \cline{2-6} 
 & intfloat/multilingual-e5-base & 5.92 & 5.92 & 6.17 & 2.12 \\ \cline{2-6} 
 & mDPR-BM35-HN1 & 3.87 & 3.64 & 4.17 & 2.11 \\ \cline{2-6} 
 & GPT2 & 1.88 & 1.55 & 1.14 & 0.88 \\ \hline
\end{tabular}%
}
\caption{Retrieval performance metrics for $k = 3$. The table presents NDCG@3, MAP@3, Recall@3, and Precision@3 across various retrieval models within the Query Translation (QT), Document Translation (DT), Direct Retrieve (DR), and Zero-shot frameworks. This comprehensive summary illustrates the effectiveness of each model in handling different aspects of cross-lingual information retrieval.}
\label{table:retrieval-metrics-k3}
\end{table*}

\begin{table*}[h]
\centering
\resizebox{\textwidth}{!}{%
\begin{tabular}{|l|l|c|c|c|c|}
\hline
\textbf{Framework} & \textbf{Model} & \textbf{NDCG@5} & \textbf{MAP@5} & \textbf{Recall@5} & \textbf{Precision@5} \\ \hline
Zero-shot & monolingual-ColBERT\_eng\_mean\_last\_hidden\_state & 19.05 & 17.11 & 24.93 & 4.99 \\ \cline{2-6} 
 & monolingual-ColBERT\_eng\_CLS\_embedding & 18.44 & 10.15 & 25.64 & 5.13 \\ \cline{2-6} 
 & xlm-roberta-base & 3.21 & 1.99 & 3.89 & 1.01 \\ \cline{2-6} 
 & intfloat/multilingual-e5-base & 6.94 & 5.87 & 10.20 & 2.04 \\ \cline{2-6}
 & monolingual-Contriever\_eng & 27.88 & 15.98 & 37.23 & 7.45 \\ \hline
QT & BM25 & 5.94 & 4.46 & 7.22 & 1.44 \\ \cline{2-6} 
 & monolingual-xlm-roberta-base\_sanskrit & 0.94 & 0.40 & 1.78 & 0.36 \\ \hline
DT & BM25 & \textbf{59.76} & \textbf{50.25} & \textbf{66.20} & \textbf{13.24} \\ \cline{2-6} 
 & ColBERT-fine-tuned\_DOT & 37.52 & 32.27 & 49.24 & 11.12 \\ \cline{2-6} 
 & contriever - fine-tuned\_DOT & 38.29 & 33.93 & 51.42 & 10.28 \\ \cline{2-6} 
 & GPT2 & 1.95 & 1.75 & 1.57 & 0.62 \\ \cline{2-6} 
 & REPLUG LSR: Mistral-7B-Instruct-v0.2 and GPT2-sup & 23.10 & 20.60 & 30.63 & 6.13 \\ \hline
DR & Fine-tuned xlm-roberta-base\_DOT & 3.98 & 3.16 & 3.88 & 1.38 \\ \cline{2-6} 
 & Fine-tuned xlm-roberta-base\_CONCAT & 3.19 & 2.71 & 2.88 & 1.30 \\ \cline{2-6} 
 & intfloat/multilingual-e5-base & 8.86 & 7.21 & 12.16 & 2.62 \\ \cline{2-6} 
 & mDPR-BM35-HN1 & 4.22 & 3.95 & 4.06 & 1.41 \\ \cline{2-6} 
 & GPT2 & 2.09 & 1.62 & 2.66 & 0.46 \\ \hline
\end{tabular}%
}
\caption{Retrieval performance metrics for $k = 5$. The table presents NDCG@5, MAP@5, Recall@5, and Precision@5 across various retrieval models within the Query Translation (QT), Document Translation (DT), Direct Retrieve (DR), and Zero-shot frameworks. This comprehensive summary illustrates the effectiveness of each model in handling different aspects of cross-lingual information retrieval.}
\label{table:retrieval-metrics-k5}
\end{table*}

\begin{table*}[h]
\centering
\resizebox{\textwidth}{!}{%
\begin{tabular}{|l|l|c|c|c|c|}
\hline
\textbf{Framework} & \textbf{Model} & \textbf{NDCG@10} & \textbf{MAP@10} & \textbf{Recall@10} & \textbf{Precision@10} \\ \hline
Zero-shot & monolingual-ColBERT\_eng\_mean\_last\_hidden\_state & 21.77 & 18.23 & 33.33 & 3.33 \\ \cline{2-6} 
 & monolingual-ColBERT\_eng\_CLS\_embedding & 20.84 & 10.84 & 33.14 & 3.31 \\ \cline{2-6} 
 & xlm-roberta-base & 3.77 & 2.22 & 4.62 & 0.46 \\ \cline{2-6} 
 & intfloat/multilingual-e5-base & 8.75 & 6.61 & 15.87 & 1.59 \\ \cline{2-6}
 & monolingual-Contriever\_eng & 30.48 & 16.67 & 45.30 & 4.53 \\ \hline
QT & BM25 & 6.86 & 4.81 & 9.90 & 0.99 \\ \cline{2-6} 
 & monolingual-xlm-roberta-base\_sanskrit & 1.22 & 0.71 & 2.99 & 0.30 \\ \hline
DT & BM25 & \textbf{62.46} & \textbf{51.30} & \textbf{74.03} & \textbf{7.40} \\ \cline{2-6} 
 & ColBERT-fine-tuned\_DOT & 40.78 & 34.33 & 60.62 & 6.61 \\ \cline{2-6} 
 & contriever - fine-tuned\_DOT & 41.70 & 35.37 & 61.82 & 6.18 \\ \cline{2-6} 
 & GPT2 & 2.09 & 2.82 & 2.99 & 0.30 \\ \cline{2-6} 
 & REPLUG LSR: Mistral-7B-Instruct-v0.2 and GPT2-sup & 26.26 & 21.89 & 40.50 & 4.05 \\ \hline
DR & Fine-tuned xlm-roberta-base\_DOT & 4.26 & 4.79 & 7.13 & 0.71 \\ \cline{2-6} 
 & Fine-tuned xlm-roberta-base\_CONCAT & 3.53 & 2.92 & 5.60 & 0.56 \\ \cline{2-6} 
  & intfloat/multilingual-e5-base & 10.74 & 8.86 & 18.26 & 1.83 \\ \cline{2-6} 
 & mDPR-BM35-HN1 & 4.74 & 4.17 & 7.67 & 0.77 \\ \cline{2-6} 
 & GPT2 & 2.54 & 1.76 & 4.01 & 0.40 \\ \hline
\end{tabular}%
}
\caption{Retrieval performance metrics for $k = 10$. The table presents NDCG@10, MAP@10, Recall@10, and Precision@10 across various retrieval models within the Query Translation (QT), Document Translation (DT), Direct Retrieve (DR), and Zero-shot frameworks. This comprehensive summary illustrates the effectiveness of each model in handling different aspects of cross-lingual information retrieval.}
\label{table:retrieval-metrics-k10}
\end{table*} 

\section{Discussion}
\subsection{Performance Evaluation}

The tables referenced as \ref{table:retrieval-metrics-k1}, \ref{table:retrieval-metrics-k3}, \ref{table:retrieval-metrics-k5}, and \ref{table:retrieval-metrics-k10} clearly display how various models performed at different cutoff points (k-values) during our tests. Remarkably, the Document Translation (DT) framework consistently showed superior performance across most metrics, largely because it uses well-developed models like BM25. These models are particularly effective in managing the complexities of translated text retrieval. Within this framework, BM25 was notably efficient, with its performance improving at higher k-values—from 40.64\% in NDCG at $k=1$ to 62.46\% at $k=10$. This indicates its strong capability in retrieving a wider array of relevant documents. Other specialized models like \textit{ColBERT} and \textit{contriever}, fine-tuned using the \textit{DOT} method, also demonstrated robust performance, reinforcing the value of tailored models in this setting.

In the Zero-shot framework, which does not rely on fine-tuning, the \textit{monolingual-Contriever\_eng} model stood out. It effectively increased its NDCG score from 18.33\% at $k=1$ to 30.48\% at $k=10$, showcasing its ability to retrieve relevant documents across broader contexts. This model excelled particularly in achieving high recall rates, proving it can find relevant documents reliably.

Performance was more mixed in the Query Translation (QT) and Direct Retrieve (DR) frameworks. In these frameworks, models like the fine-tuned \textit{xlm-roberta-base}, in various configurations, showed moderate success. However, the GPT-2 model, used in the DR framework, did not perform as well compared to more focused retrieval models. Its general-purpose design may not be as effective for specific retrieval tasks.

Zero-shot models, which operate without supervised fine-tuning, generally started strong, especially in initial retrieval accuracy. However, they tended to lag behind the fine-tuned models of the DT and DR frameworks on metrics like precision and NDCG. This highlights the critical importance of choosing the right model based on the specific needs of the retrieval task and the dataset's characteristics. The results emphasize that each model and framework brings different strengths and weaknesses, guiding us to select the most effective models for specific applications in cross-lingual information retrieval.

\subsection{Assessment of Retrieval Approaches and Model Adaptations}
Our study critically examined various retrieval approaches and their adaptability to cross-lingual settings. In the Direct Retrieve (DR) framework, which uses models like mDPR and xlm-roberta-base, the focus was on embedding-based retrieval methods that bypass traditional translation processes and aim to capture semantic similarities directly. Despite the theoretical benefits, practical implementation faced challenges, particularly achieving high precision at lower k-values, underscoring the need for more refined embedding strategies and training methods.

Furthermore, within the Document Translation (DT) framework, the adaptation of models like REPLUG LSR offered valuable insights. This model effectively used language model supervision to align retrieval processes with language predictions, enhancing the relevance and contextual accuracy of the documents it retrieved. These results suggest promising directions for integrating advanced language understanding in retrieval systems.

\section{Conclusion and Future Work}

Throughout our comprehensive evaluation of cross-lingual information retrieval (CLIR) systems, we have uncovered significant variations in performance, which primarily arise from differences in retrieval frameworks and the capabilities of the underlying models. The Document Translation (DT) framework emerged as a standout, demonstrating consistently high performance across various metrics. This success can be attributed to the effective use of the BM25 algorithm and finely tuned contriever models, which excel in understanding the nuances of English documents post-translation. In contrast, the Direct Retrieve (DR) framework, which innovatively bypasses traditional language barriers through models like mDPR and xlm-roberta-base, sometimes struggled to achieve the precision and recall rates of its more traditional counterparts. Remarkably, zero-shot models, especially the monolingual-eng-contriever, have shown great promise in handling retrieval tasks without prior fine-tuning, underscoring the potential of pre-trained models for swift deployment in diverse linguistic settings. However, the observed variability in performance across different evaluation metrics at various k-values indicates a pressing need for further optimization to boost consistency and reliability.

As we look to the future, our strategy includes expanding our dataset to encompass a broader spectrum of documents and queries. This expansion is critical for a deeper understanding of the scalability and adaptability of our existing models. We are particularly excited about integrating REPLUG \cite{shi2023replug}, a cutting-edge retrieval-augmented language modeling framework. REPLUG enhances a black-box language model with a tunable retrieval component, prepending retrieved documents to the model's input. This novel approach not only boosts the language model's performance but also allows it to guide the retrieval process effectively.

Further, we plan to implement Constraint Translation Candidates \cite{bi2020constraint}, which refines neural query translation by focusing the target vocabulary on key terms derived from the search index, thereby enhancing the relevance and accuracy of the translation outputs. Additionally, our efforts will extend to incorporating Hierarchical Knowledge Enhancement \cite{zhang2022mind}, leveraging a multilingual knowledge graph to bridge linguistic gaps in CLIR tasks. This technique promises to enrich query representations by integrating contextual knowledge from entities and their networks, smoothing over language discrepancies in the retrieval process.

Another exciting avenue is to improve our translation systems by selecting and utilizing top-K sentences from translations, which could be employed to refine retrieval effectiveness further. These planned advancements will facilitate dynamic adjustments in retrieval strategies, tailored to the linguistic characteristics of both queries and documents, thereby significantly elevating the effectiveness of CLIR systems in handling the linguistic diversity encountered in real-world applications.

Additionally, to comprehensively enhance our dataset, we will adopt methodologies from CLIRMatrix \cite{ogundepo2022africlirmatrix}, which utilizes a vast collection of bilingual and multilingual datasets extracted from Wikipedia. This approach will allow us to methodically expand our dataset with a diverse array of document and query pairs across multiple languages, fine-tuning our retrieval models to operate more effectively across the varied landscape of global languages.

\section{Limitations}
Our investigation into cross-lingual information retrieval (CLIR) systems has surfaced several significant limitations that merit further discussion. Primarily, the efficacy of both the Document Translation (DT) and Query Translation (QT) frameworks hinges critically on the accuracy of the translation mechanisms employed, such as Google Translate. Any errors introduced during translation can cascade through the retrieval process, potentially diminishing the overall effectiveness of these systems.

Moreover, our Direct Retrieve (DR) framework, designed to sidestep the translation requirement, depends heavily on the capability of models like mDPR to semantically align multilingual texts. This alignment is inherently challenging due to the intricate nuances and subtleties of language, which may not be fully captured by current models.

Additionally, the zero-shot models employed in our study, despite their robustness, displayed fluctuating performance across different languages and queries. Particularly, these models often underperform when dealing with languages or dialects that were minimally represented in their training datasets, underscoring a significant limitation in the existing pre-trained models to universally handle linguistic diversity.

A specific challenge in our research is the absence of publicly available datasets for CLIR tasks involving English queries to Sanskrit documents. This gap significantly limits the development and testing of retrieval systems aimed at accessing a rich heritage of Sanskrit scriptures. There is a pressing need to develop strategies for compiling comprehensive datasets that include diverse Sanskrit documents and corresponding queries in English to better support research in this area.

Another notable limitation of our research is the lack of in-depth exploration into the effects of varying data formats and linguistic styles—such as informal versus formal language—on retrieval accuracy. These stylistic variations play a crucial role in the practical application of CLIR systems and need comprehensive analysis to ensure these systems can adapt effectively to real-world data, which often contains a blend of many different linguistic styles.

In future studies, addressing these limitations will be crucial for advancing the adaptability and accuracy of CLIR systems, making them more effective across a broader spectrum of languages and more reflective of the varied and complex nature of human language in everyday use.

\section{Acknowledgment}
The work was supported in part by the National Language Translation Mission (NLTM): Bhashini project by Government of India. \\
We gratefully acknowledge Sai Gokul V from the Indian Institute of Technology, Kharagpur, for his diligent efforts and valuable contributions to the annotation of the dataset.

\bibliographystyle{acl}
\bibliography{custom}

% \appendix

% \section{Appendix}
% \label{sec:appendix}

\end{document}